\documentclass[10pt,twocolumn,letterpaper]{article}

\usepackage[pagenumbers]{cvpr} 

\usepackage{comment}

\usepackage{tikz}
\usetikzlibrary{positioning, arrows.meta, fit, calc, backgrounds}

\definecolor{cvprblue}{rgb}{0.21,0.49,0.74}
\usepackage[pagebackref,breaklinks,colorlinks,allcolors=cvprblue]{hyperref}

\def\paperID{*****} 
\def\confName{CVPR}
\def\confYear{2026}

\title{HSEmotion Team at ABAW-10 Competition: Facial Expression Recognition, Valence-Arousal Estimation, Action Unit Detection and Fine-Grained Violence Classification}

\author{Andrey V. Savchenko\textsuperscript{1,2}\\
\textsuperscript{1}Sber AI Lab\\
\textsuperscript{2}HSE University\\
Moscow, Russia\\
{\tt\small avsavchenko@hse.ru}
\and
Kseniia Tsypliakova\\
HSE University\\
Moscow, Russia\\
{\tt\small kdtsypliakova@gmail.com}
}

\begin{document}
\maketitle
\begin{abstract}
This article presents our results for the 10th Affective Behavior Analysis in-the-Wild (ABAW) competition. For frame-wise facial emotion understanding tasks (frame-wise facial expression recognition, valence-arousal estimation, action unit detection), we propose a fast approach based on facial embedding extraction with pre-trained EfficientNet-based emotion recognition models. If the latter model's confidence exceeds a threshold, its prediction is used. Otherwise, we feed embeddings into a simple multi-layered perceptron trained on the AffWild2 dataset. Estimated class-level scores are smoothed in a sliding window of fixed size to mitigate noise in frame-wise predictions. For the fine-grained violence detection task, we examine several pre-trained architectures for frame embeddings and their aggregation for video classification. Experimental results on four tasks from the ABAW challenge demonstrate that our approach significantly improves validation metrics over existing baselines.
\end{abstract}

\section{Introduction}
\label{sec:intro}

Understanding and modeling human affect from visual (and audiovisual) signals is a pressing challenge for modern computer vision~\cite{ramaswamy2024multimodal}: accurate emotion understanding in video is foundational for applications ranging from human–computer interaction~\cite{chowdary2023deep} and mental-health monitoring to driver safety and content moderation~\cite{guo2024development,rhethika2025ai}. These problems are particularly acute in unconstrained, real-world data because affective signals are often subtle and transient, and performance must withstand occlusions, large pose and illumination variation, domain shifts, and noisy or ambiguous annotations~\cite{kuruvayil2022emotion}. At the same time, real applications demand methods that are computationally efficient, robust to class imbalance and label bias in existing datasets, and able to fuse complementary modalities while providing temporally consistent outputs. 

The series of ABAW (Affective Behavior Analysis in-the-Wild) challenges~\cite{kollias2020analysing,kollias2021analysing,kollias2022abaw,kollias2023abaw2,kollias20246th,kollias20247th,Kollias_2025_CVPR} is a popular benchmark, which forces researchers to establish new frontiers in audiovisual understanding of human affect and behavior and multi-task learning~\cite{kollias2024behaviour4all,kollias2024distribution,kollias2021affect,kollias2021distribution,savchenko2021method} for affective computing in real-world environments. The organizers of the challenge prepared several publicly available datasets, e.g., Aff-Wild2~\cite{kollias2019expression,kollias2019deep,zafeiriou2017aff}, DVD~\cite{kollias2025dvd}, and C-EXPR~\cite{kollias2023multi}, to advance progress in this area. The most popular challenges in a series of ABAW competitions are three uni-task frame-level facial emotion understanding problems on the AffWild2 dataset: expression (EXPR) classification, Valence-Arousal (VA) estimation, and Action Unit (AU) detection. The tenth edition of the ABAW competition (ABAW-10) continues studies in these directions for facial analysis, but also addresses fine-grained Violence Detection (VD), which typically requires processing the entire frame. 

\begin{figure*}[t]
 \centering
 \includegraphics[width=0.9\linewidth]{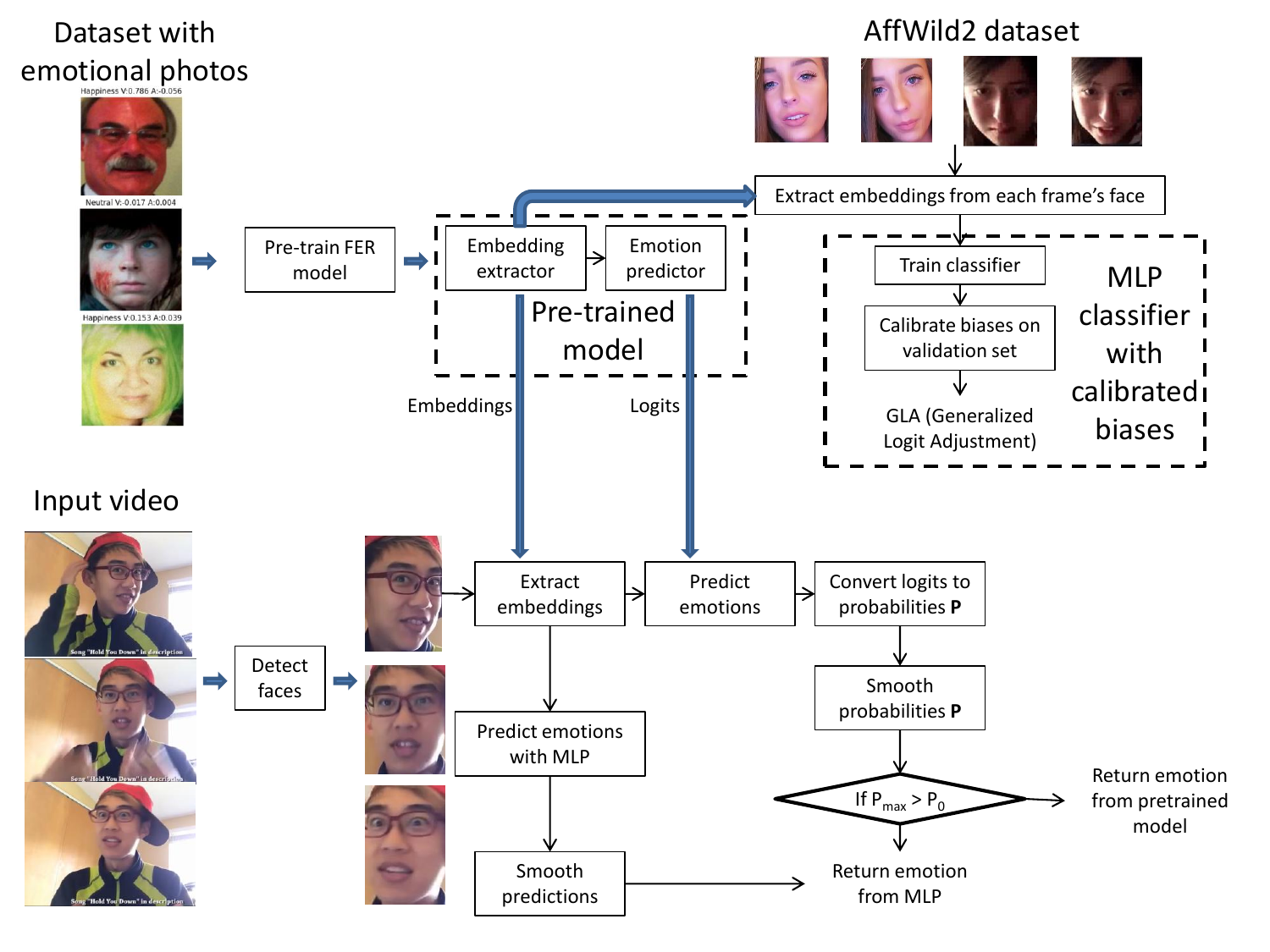}
 \caption{Proposed approach for EXPR classification.}
 \label{fig:expr_pipeline}
\end{figure*}

In this paper, we present our team's results in four tasks of the ABAW-10 competition. We are mainly focused on the most popular problem, facial expression recognition (FER), and propose the novel pipeline (Fig.~\ref{fig:expr_pipeline}) that utilizes facial embeddings extracted by EfficientNet models~\cite{tan2019efficientnet} from EmotiEffLib library~\cite{savchenko2023cvprw}, classify them with an MLP (Multi-layered Perceptron) enhanced by GLA (Generalized Logit Adjustment)~\cite{zhu2023generalized} to deal with emotion imbalance. If the pre-trained outputs emotion with very high confidence, it is used instead of the MLP's output. The frame-level predictions are smoothed, and the acoustic features may be utilized in a late-fusion (blending) setting. A similar pipeline is implemented for VA estimation and AU detection. For VD detection, we examine various visual encoders and temporal fusion techniques. We experimentally show that the proposed approach, though straightforward, is much more accurate than the baselines of the challenge organizers and also achieves high performance compared to participants' results from previous years.

\section{Related Works}\label{sec:related}
Let us consider the main lessons learned by participants of  ABAW competitions.

\textbf{EXPR Classification.} 
The most typical approach for ABAW challenges is 1) utilizing the facial emotional models, pre-trained on external datasets, for frame-level feature extraction, along with acoustic features, such as wav2vec~\cite{baevski2020wav2vec}, and 2) training the temporal deep neural network for video classification on the available training set. For example, the second and first places at ABAW-6 used Masked Autoencoders (MAE) at the first stage and Temporal Convolutional Network (TCN)~\cite{Zhou_2024_CVPR} or transformer-based feature fusion module~\cite{zhang2024effective}. The first place at ABAW-8 was achieved by using lightweight models (EfficientNet, MobileFaceNet, MobileNet, etc.) pre-trained on the AffectNet dataset~\cite{mollahosseini2017affectnet} for simultaneous prediction of facial expressions, valence and arousal~\cite{demochkina2021mobileemotiface,kollias2023abaw,kollias2019face,kollias2024behaviour4all,Savchenko_2024b_CVPR}, combined with simple averaging of predictions on top of an MLP classifier~\cite{savchenko2025leveraging}. 

A pre-trained facial encoder was used in the Video-based Noise-aware Adaptive Weighting (V-NAW)~\cite{Lee_2025_CVPR}, which addresses label ambiguity and noisy labels, spatiotemporal redundancy, and long-range dependencies in facial expressions across video frames. The third place at ABAW-8~\cite{Yu_2025_CVPR_EXPR} fuses a Residual-based Hybrid Convolutional Neural Network with audio feature expressions and frequency masking to conduct targeted processing on the original data. Semi-supervised learning (SSL)~\cite{Yu_2024_CVPR} with pseudo-label generation for unlabeled faces achieved third place in the ABAW-6 competition. 

Many recent papers also utilize multimodal techniques based on CLIP, e.g., to classify visual features using an MLP enhanced with Conditional Value at Risk~\cite{lin2024robust}. CLIP features also worked slightly better than Dual-Direction Attention Mixed Feature Network (DDAMFN)~\cite{zhang2023dual} for FER~\cite{Cabacas-Maso_2025_CVPR}. Moreover, the second place at ABAW-8~\cite{zhou2025emotion} incorporated TCN and a transformer encoder to combine fine-tuned CLIP with the Aff-Wild2 dataset.

The \textbf{VA Estimation} task requires predicting continuous emotion representations. The solutions are similar to EXPR classification, e.g., top results were obtained by participants~\cite{zhou2023continuous,savchenko2023cvprw} that used EfficientNet models pre-trained on AffectNet. Pre-trained DDAMFN for face encoding was fed into the LSTM recurrent network in~\cite{Cabacas-Maso_2025_CVPR}. Moreover, the winner of ABAW-5 used the fine-tuned MAE~\cite{zhang2023abaw5}. The third place at ABAW-8~\cite{zhou2025emotion} used a similar CLIP-based pipeline for several different tasks. However, the AffWild2 dataset with labeled valence and arousal is several times larger than the expression-label dataset, so more sophisticated fusion and fine-tuning techniques usually yield better results. Time-aware Gated Fusion framework was introduced in~\cite{Lee_2025_ICCV} to combine visual (ResNet-50) and acoustic (VGGish) encoders. Multi-modal Attention for Valence-Arousal Emotion Network (MAVEN) integrates visual, audio, and textual modalities through a bi-directional cross-modal attention mechanism~\cite{Ahire_2025_CVPR}. Gated Recursive Joint Cross Attention (GRJCA)~\cite{Rajasekhar_2025_CVPR} with two audio-visual fusion models introduced an efficient gating mechanism to effectively capture intermodal relationships by handling weakly complementary relationships between audio and visual modalities. The winner of ABAW-8~\cite{Yu_2025_CVPR_VA} integrated visual and audio information via multiscale TCNs to capture temporal dynamics, and employed cross-modal attention mechanisms to effectively fuse features extracted by pre-trained audio and facial encoders, creating joint representations that capture relationships across modalities.

\textbf{AU detection} problem requires locating 12 micro-expressions. It typically uses similar techniques to pre-train embeddings on external datasets and to train multimodal temporal fusion modules on the AffWild2 dataset. The third place at ABAW-8 was obtained by EfficientNet's embeddings and AutoML techniques for multi-label classification~\cite{savchenko2025leveraging}. The aforementioned fine-tuning of CLIP took second place~\cite{zhou2025emotion}. The winner of ABAW-8 was the Spatial Transformer Network (STN), TCN, and dual-modal feature extraction (ConvNeXt on images and Whisper on audio) with a multiscale self-attention and an adaptive sliding window to capture contextual relationships~\cite{Yu_2025_CVPR_AU}.

\textbf{Video-based Violence Detection (VD)} 
has recently gained attention as a fine-grained behavior understanding task~\cite{negre2024literature}, and was first formalized at scale in the ABAW-9 Fine-Grained VD Challenge~\cite{Kollias_2025_ICCV}, which builds on the DVD dataset with frame-level violent/non-violent annotations~\cite{kollias2025dvd}. Beyond handcrafted features and early deep models such as shallow or 3D CNN-based architectures for surveillance violence recognition~\cite{dundar2024shallow}, recent work has explored multimodal and attention-based pipelines, including audio–visual fusion~\cite{azim2026explainable,nardelli2024josenet} and hyperbolic-space representations for weakly supervised violence localization~\cite{zhou2024learning}. The ABAW-9 winning solution~\cite{Kim_2025_ICCV} adopts this trend by proposing a multimodal framework that integrates motion features from VideoPrism, static spatial cues from CLIP, and semantic information from caption-derived text embeddings, which are encoded with GRU-based sequence models and fused via a two-stage cross-modal attention mechanism.

\section{Proposed Approach}\label{sec:proposed}

For all ABAW-10 tasks involving facial analysis (EXPR, VA, and AU), we use lightweight architectures (EfficientNet, DDAMFN, MobileViT, etc.)~\cite{Savchenko_2024a_CVPR} from the EmotiEffLib library (former HSEmotion~\cite{savchenko2022hsemotion}). They are pre-trained to recognize 8 basic expressions (and, for some models, valence and arousal) from static photos on the AffectNet dataset~\cite{mollahosseini2017affectnet}. The resulting models (EmotiEffNet-B0, MT-EmotiEffNet, MT-DDAMFN, etc.) can be used to predict facial expressions and extract emotional embeddings that can be utilized in various downstream affective computing problems. As our pre-trained models work with rather high-resolution facial images (typically 224x224), we used cropped facial images from the AffWild2 dataset, which was officially provided by the challenge organizers. 

\subsection{Frame-level Expression recognition}
Fig.~\ref{fig:expr_pipeline} illustrates the main parts of the proposed pipeline for FER in video. In the EXPR classification task of the ABAW-10 challenge, each video frame should be associated with one of $C=8$ emotions (Neutral, Anger, Disgust, Fear, Happiness, Sadness, Surprise, Other). The training set contains 248 videos ($N=585,317$ labeled frames in total), and the validation set consists of 70 videos (280,532 labeled frames). By using EmotiEffNet, we extract embeddings $\mathbf{x}_n$ for facial region of every $n$-th frame from the training set (with known label $y_n \in \{1,...,C\}$, and train an MLP with 1 hidden layer and $C$ outputs using weighted softmax loss:
\begin{align}
&\mathcal{L}_{EXPR}(\mathbf{x}_n,y_n)=\nonumber\\
&=-\log softmax(z(\mathbf{x}_n)_{y_n}) \cdot \underset{y \in \{1,...,C\}} \max \frac{N_{y_n}}{N},
\end{align}
where $N_{y_n}$ is the total number of training examples of the $y_n$-th expression, $z(\mathbf{x})$ is the $C$-dimensional vector of logits at the output of the last MLP's layer.

Next, since the training set in AffWild2 is highly imbalanced, we use GLA~\cite{zhu2023generalized} to calibrate biases in the last layer. We compute the scores (emotional class posterior probabilities) at the output of MLP ($p_y^{MLP}=softmax(z(\mathbf{x})_y)$) for a validation set, and look for biases $\mathbf{b}^*=[b_1^*,...,b_C^*]$ that maximize F1 score for adjusted logits $p_y \cdot b_y^*$ (using logarithmic scale). In particular, we initialize the biases by the class prior probabilities ($b_y=N_y/N$), and perform the coordinate search in a grid $[-2;2]$.

During inference, we detect faces in each $t$-th frame of the input video, feed the cropped face into pre-trained EmotiEffNet, extract embeddings $\mathbf{x}(t)$ and scores $\mathbf{p}^{Pre-trained}(t)$ at the output of its penultimate and last layers, respectively. We consider a sliding window of $T+1$ frames ($t-T/2,..., t, ..., t+T/2$, and smooth the predictions to remove frame-level noise by computing average emotional probabilities $\tilde{\mathbf{p}}^{Pre-trained}(t)$~\cite{savchenko2023cvprw}. Next, we propose filtering the results based on the pre-trained model's confident predictions: we map them from AffectNet to AffWild2 classes and select the class with the highest probability. If
\begin{equation}
\underset{y \in \{1,...,C\}} \max \tilde{p}_y^{Pre-trained}(t) > p_0,
\end{equation}
and the class that corresponds to this maximal probability is present in AffWild2 (i.e., it is not Contempt), we simply predict this class for the $t$-th frame. Here, $p_0$ is a tunable threshold (typically in a range 0.8-0.9).

Otherwise, we feed $\mathbf{x}(t)$ into MLP with adjusted logits
\begin{equation}
\mathbf{p}^{MLP}(t)=softmax(\mathbf{z}(\mathbf{x}(t))) \cdot \mathbf{b}^*,
\end{equation}
and average these predictions in a sliding window, and return the class that corresponds to the maximal mean probability $\tilde{\mathbf{p}}^{MLP}(t)$.


Additionally, we consider using information from the audio modality. In particular, we extract wav2vec 2.0 features~\cite{baevski2020wav2vec} for audio segments and scale them to the scale of video frames. Next, we train separate MLPs, and the outputs of MLPs for facial and audio modalities are blended with weight $w$:
\begin{equation}
\mathbf{p}(t)=w\cdot \mathbf{p}^{MLP}(t) + (1-w)\cdot \mathbf{p}^{audio}(t).
\end{equation}

\subsection{VA Estimation and AU detection}
The pipeline described in the previous subsection can be adapted for other tasks on the AffWild2 dataset. For VA estimation, it is required to predict valence $V$ and arousal $A$ (both in a range $[-1;1]$) for each frame of an input video, given the training (356 videos, 1653710 labeled frames) and Validation (76 videos, 376,323 frames in total) sets. It was experimentally found that, in this setting, the MT-DDAMFN pre-trained model~\cite{Savchenko_2024a_CVPR} produces the best results. Thus, we use them to train an MLP without a hidden layer using the loss that combines MSE (Mean Squared Error) and CCC (Concordance Correlation Coefficient):
\begin{align}
&\mathcal{L}_{VA}(\mathbf{x}_n,y_n^{(V)},y_n^{(A)})= \nonumber \\
& =(z_V(\mathbf{x}_n)-y_n^{(V)})^2 +(z_A(\mathbf{x}_n)-y_n^{(A)})^2 -\\
& -(CCC(z_V(\mathbf{x}_n),y_n^{(V)})+CCC(z_A(\mathbf{x}_n),y_n^{(A)})).\nonumber
\end{align}

The AU detection challenge is a multi-label frame-level classification task: each frame is associated with 12 AUs. The splits are: 295 videos in the training set (1,356,694 frames) and 105 videos in the Validation set (445,836 frames). For each frame, a binary label is available for each AU. We use the same embeddings of EmotiEffLib for feature extraction as in the EXPR classification challenge, and train an MLP with 1 hidden layer and 12 outputs with sigmoid activation $\sigma(\cdot)$ on top of facial embeddings using weighted BCEWithLogitsLoss:
\begin{align}
&\mathcal{L}_{AU}(\mathbf{x}_n,\mathbf{y}_n)=-\sum\limits_{c=1}^{12}{\left(w_c^{pos} \cdot y_{n;c}\log\sigma(z(\mathbf{x}_n)_c)+\right.}\nonumber\\
& \left. + (1-y_{n;c})\log(1-\sigma(z(\mathbf{x}_n)_c)\right),
\end{align}
where positive class weights $w_c^{pos}$ are set to be the ratio of the total number of training frames with AU $c$ relative to the total size of the training set.

In addition, we noticed that the classifier trained on logits (outputs of the last layer) of a pre-trained model shows rather high accuracy, so we also examined the blending of two MLPs trained for facial embeddings and logits.

Finally, it is important to properly choose the thresholds for the sigmoid outputs of the trained MLP. We examine two cases: default value (0.5) for each of 12 AUs, and adjusted search for the best threshold (from a range $[0.1, 0.2, ..., 0.9]$, which optimizes validation F1-score for each AU.

\subsection{Fine-Grained Violence Detection}

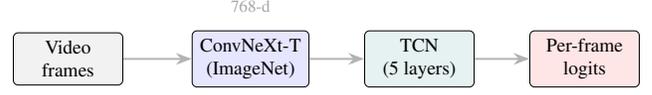
\begin{figure}[t]
\centering
\resizebox{\columnwidth}{!}{%
\begin{tikzpicture}[
    >=Stealth,
    node distance=0.5cm and 0.7cm,
    box/.style={draw, rounded corners=2pt, minimum height=0.6cm, minimum width=1.6cm,
                font=\footnotesize, align=center, fill=#1!10},
    box/.default=blue,
    arr/.style={->, thick, color=gray!60},
    lbl/.style={font=\scriptsize, color=gray!70},
]
\node[box=gray] (vid) {Video\\frames};
\node[box=blue, right=1.0cm of vid] (cnn) {ConvNeXt-T\\(ImageNet)};
\node[box=teal, right=0.8cm of cnn] (tcn) {TCN\\(5 layers)};
\node[box=red, right=0.8cm of tcn] (out) {Per-frame\\logits};
\node[lbl, above=0.15cm of cnn] {768-d};
\draw[arr] (vid) -- (cnn);
\draw[arr] (cnn) -- (tcn);
\draw[arr] (tcn) -- (out);
\end{tikzpicture}%
}
\caption{Our best single-stream frame-level violence detection pipeline. ConvNeXt-T (pretrained on ImageNet-1K) extracts 768-d per-frame features, processed by a 5-layer dilated TCN.}
\label{fig:vd_pipeline}
\end{figure}

\begin{table*}[t]
 \centering
 \begin{tabular}{ccccc}
 \toprule
 Method & Modality & Is ensemble? & F1-score $P_{EXPR}$ & Accuracy \\
 \midrule
 Baseline VGGFACE (MixAugment)~\cite{Kollias_2025_CVPR} & Faces & No & 25.0 & -\\
\hline
CLIP+LSTM~\cite{Cabacas-Maso_2025_CVPR} & Video & Yes & 34.1 & - \\
CLIP~\cite{lin2024robust} & Faces & No & 36.0 & - \\ 

EfficientNet-B0~\cite{savchenko2022cvprw} & Faces & No & 40.2 & - \\
V-NAW~\cite{Lee_2025_CVPR} & Video & No & 41.81 & - \\
SSL + Temporal+ Post-process~\cite{Yu_2024_CVPR}& Audio/video & Yes & 44.43 & - \\
Best EfficientNet + wav2vec~\cite{savchenko2025leveraging} &Audio/video & Yes &  44.59 & 55.32 \\
CLIP+TCN~\cite{zhou2025emotion} & Video & Yes &  46.51 & - \\
MAE+Transformer feature fusion~\cite{zhang2024effective} & Audio/video & Yes & 55.55 & - \\ 
\hline
wav2vec 2.0, focal loss & Audio & No & 30.50 & 40.08 \\
wav2vec 2.0, focal loss, GLA & Audio & No & 32.33 & 42.15 \\
wav2vec 2.0, focal loss, GLA, smoothing & Audio & No & 38.75 & 51.74 \\
EmotiEffNet& Faces & No & 38.68 & 47.59 \\
EmotiEffNet, GLA& Faces & No & 41.40 & 51.62 \\
EmotiEffNet, GLA + filtering& Faces & No & 42.50 & 52.22 \\
EmotiEffNet, focal loss& Faces & No & 38.99 & 50.22 \\
EmotiEffNet, focal loss, GLA& Faces & No & 40.18 & 51.90 \\
wav2vec 2.0+EmotiEffNet, GLA& Audio/video & Yes & 43.11 & 53.40 \\
EmotiEffNet, GLA, smoothing & Faces & No & 44.85 & 55.41 \\
EmotiEffNet, GLA, filtering + smoothing & Faces & No & 45.79 &55.69 \\
EmotiEffNet + wav2vec 2.0,  GLA, smoothing & Audio/video & No & 46.04 & 57.50 \\
EmotiEffNet + wav2vec 2.0, GLA, filtering + smoothing & Audio/video & No & 47.40 & 57.98 \\
 \bottomrule
 \end{tabular}
 \caption{Main results of Expression Recognition for the Aff-Wild2’s validation set.}
 \label{tab:expr}
\end{table*}

DVD~\cite{kollias2025dvd} is a dataset for video-based violence detection, where each frame must be classified as violent or non-violent. The training set consists of 103 videos ($\approx$772K annotated frames, of which 44\% are violent), and the official validation set contains 17 videos ($\approx$77K frames). Unlike the facial analysis tasks above, VD requires analyzing the full frame to capture body motion, person interactions, and scene context. 

Each video is split into clips of $L{=}32$ frames with a temporal stride (frame step) of 2, resized to $224 \times 224$ with ImageNet normalization. We examine a range of 2D and 3D backbones, temporal heads, and multimodal fusion strategies; the full comparison is presented in Section~\ref{sec:exper}. Below, we describe the best single-stream configuration and the best multimodal variant.

The best single-stream model (Fig.~\ref{fig:vd_pipeline}) uses a ConvNeXt-T~\cite{liu2022convnet} backbone pretrained on ImageNet-1K~\cite{deng2009imagenet} to extract 768-dimensional per-frame features, followed by a 5-layer dilated TCN~\cite{bai2018empirical}. The model is trained with AdamW, differential learning rates (backbone at $0.15\times$ the base rate), OneCycleLR scheduling, and TrivialAugmentWide~\cite{muller2021trivialaugment} applied independently to each frame. We also examined several additional modifications, including transformer-based temporal heads, boundary-aware losses, and structured temporal decoding, but none improved upon the strongest ConvNeXt-T baselines in our setting.

The best multimodal variant augments this RGB stream with skeleton features extracted by MediaPipe Pose~\cite{lugaresi2019mediapipe}: top-2 poses per frame, their keypoint coordinates, inter-frame velocities, and pairwise interaction distances (total 406 dims), projected to 256 dims and fused with RGB via cross-attention. The fused representation is processed by a BiLSTM. Here, the ConvNeXt encoder is first fine-tuned on DVD in the single-stream setting, then loaded as a frozen backbone, while the fusion module and BiLSTM are trained from scratch.

Both models use weighted cross-entropy with the positive (violent) class weight scaled by 1.15 to compensate for class imbalance. During inference, a sliding window with a stride of 16 is applied over the full video, per-frame probabilities are averaged across overlapping windows, and the final prediction is obtained by thresholding at 0.5.

\section{Experimental Results}\label{sec:exper}

\subsection{Facial Expression Recognition}
To evaluate our approach on the EXPR classification task, we tested various models on the Aff-Wild2 validation set. Table~\ref{tab:expr} summarizes the macro-averaged F1-score $P_{EXPR}$(official metric of organizers) and accuracy achieved by different methods. The baseline ResNet-50 model achieves relatively modest performance, highlighting the difficulty of expression recognition in unconstrained, in-the-wild settings. More advanced architectures that incorporate temporal modeling or multimodal fusion (e.g., LSTM-based or transformer-style approaches) improve the results but still remain limited by class imbalance, noisy frame-level annotations, and unstable frame-wise predictions.

Our approach significantly improves the validation performance by combining strong pre-trained facial emotion encoders with a lightweight classification pipeline. In particular, embeddings extracted from EfficientNet-based emotional models provide discriminative frame-level representations that are further refined by a simple MLP classifier trained on the Aff-Wild2 data. The use of logit adjustment helps mitigate the severe imbalance across expression classes, while confidence-based filtering of predictions from the pre-trained model enables the system to directly exploit highly reliable outputs. Finally, temporal smoothing over a sliding window reduces prediction noise across adjacent frames, leading to more stable video-level behavior. As a result, the proposed pipeline achieves substantially higher F1 scores and accuracy than the challenge baseline and demonstrates competitive performance relative to more complex temporal architectures, despite relying on a much simpler, computationally efficient design.

\subsection{Valence-Arousal Estimation}

\begin{table}[t]
 \centering
 \begin{tabular}{p{4.1cm}cccc}
 \toprule
Method  & CCC\_V & CCC\_A& $P_{VA}$ \\
 \midrule
 Baseline ResNet-50~\cite{Kollias_2025_CVPR} & 0.24 & 0.20 & 0.22\\
 \hline
MAVEN~\cite{Ahire_2025_CVPR} & 0.307 & 0.305 & 0.3061 \\
DDAMFN+LSTM~\cite{Cabacas-Maso_2025_CVPR} & - & - & 0.479 \\
Best EfficientNet~\cite{Savchenko_2024a_CVPR} & 0.490 & 0.596 & 0.543 \\
Time-aware Gated Fusion~\cite{Lee_2025_ICCV} & 0.427 & 0.676 & 0.552 \\
GRJCA~\cite{Rajasekhar_2025_CVPR} & 0.459 & 0.652 & 0.555\\
MultiScaleTCN~\cite{Yu_2025_CVPR_VA} & 0.467 & 0.663 & 0.565\\
CLIP+TCN~\cite{zhou2025emotion} & 0.562 & 0.612 & 0.587 \\
 \hline
wav2vec 2.0 & -0.011 & 0.244 & 0.116 \\
LibreFace~\cite{chang2024libreface}, MLP & 0.307 & 0.418 & 0.362 \\
MT-DDAMFN, pretrained & 0.413 & 0.229 & 0.321 \\
MT-DDAMFN, MLP & 0.469 & 0.538 & 0.503 \\
MT-DDAMFN, MLP, smoothing & 0.510 & 0.615 & 0.562 \\

 \bottomrule
 \end{tabular}
 \caption{Main results of Valence-Arousal Estimation for the Aff-Wild2’s validation set.}
 \label{tab:va}
\end{table}

For valence–arousal (VA) estimation, Table~\ref{tab:va} reports CCC for predicted valence (CCC\_V), arousal (CCC\_A) and their mean $P_{VA}=\frac{CCC\_V+CCC\_A}{2}$ (official metric of challenge organizers). Our experiments confirm that leveraging strong pre-trained face encoders and simple post-processing yields competitive continuous emotion predictions: the MLP trained on MT-DDAMFN~\cite{Savchenko_2024a_CVPR} logits with sliding-window smoothing attains CCC\_V = 0.510 and CCC\_A = 0.615, substantially outperforming the ResNet-50 baseline (0.24 / 0.20) and improving on vanilla EfficientNet encoders (0.490 / 0.596). The results also show that different design choices trade off Valence vs Arousal performance, e.g., time-aware fusion methods reach higher arousal CCC (0.676) but lower valence, and that purely audio systems (wav2vec 2.0) perform poorly for valence (negative CCC) unless tightly integrated with visual features. Overall, smoothing and modest multimodal fusion provide most of the practical gain on Aff-Wild2: our lightweight pipeline narrows the gap to heavier transformer/TCN-based fusion approaches while remaining computationally efficient and easy to reproduce.

\subsection{Action Unit Detection}

\begin{table}[t]
 \centering
 \begin{tabular}{p{6cm}c}
 \toprule
 Method & F1-score $P_{AU}$ \\
 \midrule
 Baseline VGGFACE~\cite{Kollias_2025_CVPR} & 39.0\\
\hline
CLIP~\cite{lin2024robust} & 43.0\\
DDAMFN+LSTM, threshold 0.5~\cite{Cabacas-Maso_2025_CVPR} & 41.1 \\
DDAMFN+LSTM, best thresholds~\cite{Cabacas-Maso_2025_CVPR} & 45.1 \\
AUD-TGN~\cite{Yu_2024_AUD} & 53.7\\
Best EfficientNet~\cite{savchenko2025leveraging} & 54.5 \\
ConvNeXt + Whisper+STN +TCN + Sliding Window~\cite{Yu_2025_CVPR_AU} & 56.1 \\
MAE+Transformer Fusion~\cite{zhang2024effective} & 56.9 \\
MAE ViT + TCN~\cite{Zhou_2024_CVPR} & 57.6\\
CLIP+TCN~\cite{zhou2025emotion} & 58.0 \\
\hline
wav2vec 2.0 & 31.3 \\
EmotiEffNet, logits, threshold 0.5 & 47.5 \\
EmotiEffNet, logits, best thresholds & 48.3 \\
EmotiEffNet, logits, smoothing, threshold 0.5 & 48.6 \\
EmotiEffNet, logits, smoothing, best thresholds & 49.3 \\
EmotiEffNet, embeddings, threshold 0.5 & 52.7 \\
EmotiEffNet, embeddings, best thresholds & 52.9 \\
EmotiEffNet, embeddings, smoothing, threshold 0.5 & 54.0 \\
EmotiEffNet, embeddings, smoothing, best thresholds & 54.2 \\
EmotiEffNet, logits + embeddings, threshold 0.5 & 53.7 \\
EmotiEffNet, logits + embeddings, best thresholds & 53.9 \\
EmotiEffNet, logits + embeddings, smoothing, threshold 0.5 & 54.6 \\
EmotiEffNet, logits + embeddings, smoothing, best thresholds & 54.7 \\
 \bottomrule
 \end{tabular}
 \caption{Main results of Action Unit Detection for the Aff-Wild2’s validation set.}
 \label{tab:au}
\end{table}

For AU detection, macro-averaged F1-score $P_{AU}$ is shown in Table~\ref{tab:au}. 
These results demonstrate a clear progression from simple baselines to heavier multimodal, temporal fusion methods: the VGGFACE baseline scores only 39.0\% F1, while our EmotiEffNet-based classifiers (logits, embeddings, with smoothing and threshold tuning) consistently push performance into the 54.7\% F1, closing a large part of the gap to top pipelines (54–56\% F1) and outperforming several audio-only or logits-only solutions. At the same time, state-of-the-art fusion approaches that combine strong visual pretraining with temporal models and audio report higher scores, indicating there remains headroom for gains from richer temporal and audio integration. Overall, our lightweight, calibration-aware AU system, augmented by embedding/logit blending and per-AU weighting, offers a favorable trade-off between accuracy and computational cost, improving substantially over naive baselines while remaining much simpler and easier to reproduce than the top fusion architectures.

\subsection{Fine-Grained Violence Detection}

\begin{table}[t]
\centering
\setlength{\tabcolsep}{3pt}
\begin{tabular}{p{3.7cm}ccc}
\toprule
Method & $F1_V$ & $F1_{NV}$ & Macro $F1$ \\
\midrule
\multicolumn{4}{l}{\textit{Baseline}} \\
ResNet-50 + BiLSTM~\cite{Kollias_2025_ICCV} & 0.56 & 0.71 & 0.640 \\
\hline
\multicolumn{4}{l}{\textit{3D / video backbones}} \\
SlowFast-R50         & 0.303 & 0.661 & 0.482 \\
R3D-18 + Conv1d      & 0.186 & 0.797 & 0.491 \\
S3D + Conv1d         & 0.268 & 0.781 & 0.524 \\
R(2+1)D-18 + Conv1d  & 0.335 & 0.717 & 0.526 \\
VideoMAE-Crime       & 0.283 & 0.781 & 0.532 \\
VideoMAE-Small       & 0.544 & 0.659 & 0.602 \\
S3D + BiLSTM         & 0.462 & 0.752 & 0.607 \\
R(2+1)D + BiLSTM     & 0.517 & 0.778 & 0.647 \\
\hline
\multicolumn{4}{l}{\textit{2D backbone (single-stream RGB)}} \\
ResNet-18 + Conv1d      & 0.556 & 0.769 & 0.663 \\
Video Swin-T            & 0.530 & 0.812 & 0.671 \\
EfficientNet-B0 + BiLSTM& 0.681 & 0.757 & 0.719 \\
ConvNeXt-T + BiLSTM     & 0.745 & 0.820 & 0.782 \\
ConvNeXt-T + TCN        & 0.738 & 0.828 & \textbf{0.783} \\
\hline
\multicolumn{4}{l}{\textit{Two-stream: RGB + Optical Flow}} \\
TwoStr. concat + BiLSTM & 0.658 & 0.801 & 0.730 \\
TwoStr. gated + BiLSTM  & 0.669 & 0.800 & 0.735 \\
\hline
\multicolumn{4}{l}{\textit{Multi-modal: RGB + Skeleton}} \\
ConvNeXt-T + Skel. (early) & 0.692 & 0.810 & 0.751 \\
Skel. attn + TCN           & 0.723 & 0.797 & 0.760 \\
Skel. attn + BiLSTM        & 0.715 & 0.828 & 0.772 \\
\bottomrule
\end{tabular}
\caption{Main validation results and ablation study for Fine-Grained Violence Detection on the official DVD validation set. $F1_V$ and $F1_{NV}$ denote per-class F1-scores for violence and non-violence, respectively.}
\label{tab:vd}
\end{table}

Table~\ref{tab:vd} summarizes the results of our fine-grained violence detection experiments on the official DVD validation set.

We compare several groups of approaches, including 3D video backbones (e.g., Video Swin~\cite{liu2022video}, VideoMAE~\cite{tong2022videomae}, and R(2+1)D~\cite{tran2018closer}) and 2D image encoders (e.g., ResNet~\cite{he2016deep}, EfficientNet~\cite{tan2019efficientnet}, and ConvNeXt~\cite{liu2022convnet}) combined with temporal aggregation.

All 2D backbones were initialized with weights pretrained on ImageNet-1K~\cite{deng2009imagenet}, while 3D video architectures were initialized with models pretrained on the Kinetics dataset~\cite{kay2017kinetics}. 
For VideoMAE-Crime we used a publicly available checkpoint fine-tuned for violence detection. 
The official evaluation metric is the macro-averaged F1-score across the violent and non-violent classes.

The results show that stronger 2D visual encoders combined with lightweight temporal modeling outperform both the challenge baseline and the examined 3D architectures. In particular, the ResNet-50 + BiLSTM baseline from the ABAW-9 challenge achieves macro F1 $0.640$ on the official DVD validation set, while our ConvNeXt-T models provide the best performance: $0.782$ with a BiLSTM head and $0.783$ with a TCN. This suggests that ImageNet-pretrained frame encoders provide highly discriminative spatial representations, while simple temporal heads are sufficient to capture short-term temporal dynamics in DVD videos. In contrast, most 3D/video backbones perform substantially worse, with the best of them, R(2+1)D~\cite{tran2018closer} + BiLSTM, reaching only $0.647$ macro F1.

We also examined multimodal extensions. Two-stream fusion with optical flow does not improve over the RGB-only ConvNeXt-T pipeline, achieving at most $0.735$ macro F1. Skeleton-based models are more competitive: the best cross-attention skeleton variant with BiLSTM reaches $0.772$, which is close to the best RGB-only setup and improves the F1-score for the non-violent class up to $0.828$. 

Additionally, we explored several training modifications, including focal and boundary-aware losses, but did not observe consistent gains relative to the best ConvNeXt-T + TCN setup.

Among the evaluated methods, ConvNeXt-T + TCN achieved the best performance on the official DVD validation split, improving macro F1 by more than 0.14 compared to the ABAW-9 challenge baseline. This result indicates that a strong pretrained 2D backbone combined with a lightweight temporal modeling provides an effective solution for frame-level violence detection on DVD.

\section{Conclusion}
In summary, this work introduces a lightweight, calibration-aware pipeline (Fig.~\ref{fig:expr_pipeline}) that closes several practical gaps between research-grade affect recognition and deployable systems: it combines efficient single-frame feature extraction with simple temporal smoothing and modality fusion to produce temporally consistent, well-calibrated predictions while explicitly addressing class imbalance and annotation noise. The proposed design balances accuracy, robustness, and computational cost, making it suitable for real-world, in-the-wild scenarios where occlusion, pose/illumination variation, and domain shift are common. The source code to reproduce the experiments for EXPR, VA, and AU is publicly available\footnote{\url{https://github.com/sb-ai-lab/EmotiEffLib}}, while the source code for VD is available in a separate repository\footnote{\url{https://github.com/kdtsypliakova/ABAW-10-Violence-Detection}}. It is important to emphasize that, in contrast to our previous participation in ABAW competitions~\cite{Savchenko_2024a_CVPR,savchenko2025leveraging}, which used the TensorFlow 2.x framework for model training, we switched to PyTorch, making our results more suitable for practitioners and researchers. While our experiments demonstrate the benefits of this pragmatic approach, future directions include stronger self-supervised pretraining, domain-adaptive calibration, and richer temporal models to further improve long-range consistency and uncertainty quantification for safety-sensitive applications.

{
    \small
    \bibliographystyle{ieeenat_fullname}
    \bibliography{main}
}


\end{document}